\documentclass[conference]{IEEEtran}
\usepackage{cite}
\usepackage{amsmath,amssymb,amsfonts}
\usepackage{algorithmic}

\DeclareMathOperator*{\argmin}{arg\,min}
\usepackage{graphicx}
\ifCLASSOPTIONcompsoc
\usepackage[caption=false,font=normalsize,labelfon
t=sf,textfont=sf]{subfig}
\else
\usepackage[caption=false,font=footnotesize]{subfi
g}
\fi
\usepackage{textcomp}
\usepackage{xcolor}
\usepackage{stfloats}
\usepackage{gensymb}
\def\BibTeX{{\rm B\kern-.05em{\sc i\kern-.025em b}\kern-.08em
    T\kern-.1667em\lower.7ex\hbox{E}\kern-.125emX}}
\begin{document}

\title{Reinforcement Learning Based Optimal Camera Placement for Depth Observation of Indoor Scenes 
}

% \author{\IEEEauthorblockN{1\textsuperscript{st} Yichuan Chen}
% \and

\author{\IEEEauthorblockN{
    Yichuan Chen\IEEEauthorrefmark{1}, 
    Manabu Tsukada\IEEEauthorrefmark{1}, and 
    Hiroshi Esaki\IEEEauthorrefmark{1}
}
\IEEEauthorblockA{\IEEEauthorrefmark{1}
Graduate School of Information Science and Technology, The University of Tokyo, Tokyo, Japan\\
Email: \textit{nycshisan@gmail.com,~tsukada@hongo.wide.ad.jp,~hiroshi@wide.ad.jp}}
}

\maketitle

\begin{abstract} 
Exploring the most task-friendly camera setting---optimal camera placement (OCP) problem---in tasks that use multiple cameras is of great importance. However, few existing OCP solutions specialize in depth observation of indoor scenes, and most versatile solutions work offline. To this problem, an OCP online solution to depth observation of indoor scenes based on reinforcement learning is proposed in this paper. The proposed solution comprises a simulation environment that implements scene observation and reward estimation using shadow maps and an agent network containing a \textit{soft actor-critic (SAC)}-based reinforcement learning backbone and a feature extractor to extract features from the observed point cloud layer-by-layer. Comparative experiments with two state-of-the-art optimization-based offline methods are conducted. The experimental results indicate that the proposed system outperforms seven out of ten test scenes in obtaining lower depth observation error. The total error in all test scenes is also less than 90\% of the baseline ones. Therefore, the proposed system is more competent for depth camera placement in scenarios where there is no prior knowledge of the scenes or where a lower depth observation error is the main objective.
\end{abstract}


\section{Introduction}\label{chap:intro} 
Industrial automation using cyber-physical systems is being researched and developed for the next industrial revolution (Industry 4.0). These systems often require the use of automated sensors to complete the information gathering in the workplace. Depth cameras are a kind of such automated sensors. In many application scenarios of depth cameras such as surveillance or 3D reconstruction, depth information obtained by a single depth camera is insufficient for the task. Therefore, it is necessary to introduce multiple depth cameras to conduct a comprehensive observation of the observed object to overcome this limitation; however, it leads to another problem: what arrangement should be employed for these depth cameras to achieve this goal more effectively. This problem is referred to as the \textbf{optimal camera placement (OCP)} problem. The solutions for the OCP problem can be widely applied to the automatic planning of various multicamera application scenarios in Industry 4.0, which can help improve the adaptability and resource utilization of relevant systems in the working environment.

One of the main ideas for general OCP is to determine the possible solutions in the candidate camera placement pool. Each element of the candidate pool combines a set of camera parameters such as camera positions or orientations, which are obtained by discretizing the possible range of parameters. \cite{morsly2011particle} uses an idea based on particle swarm optimization (PSO), which simulates the information-sharing behavior of migrating animals to search for the optimum. Many inertial particles move and search in parallel in the camera parameter candidate pool with a tendency to the current global optimum. Since the camera parameter candidate pool is a discrete solution space, it uses a variant of the PSO algorithm called \textit{binary PSO}. Another two similar research studies focused on \cite{chrysostomou2012optimum} \textit{artificial bee colonies} and \cite{zhang2016differential} \textit{differential evolution}. Other studies that do not discretize camera parameters use greedy search algorithms to find the OCP. \cite{liu2013optimal} proposed trans-dimensional simulated annealing (TDSA) that uses a probabilistic decision model to enable the increase or decrease in the number of cameras to be two extra search directions, i.e., the dimension of the model becomes the model variable to be optimized.

In contrast to exploring general solutions, other studies focus on developing dedicated OCP solutions for specific problems. One such example is the most researched specific issue of OCP, i.e., surveillance. In contrast to the research on applying general solutions to surveillance problems, many studies address specific aspects of the surveillance problem \cite{yabuta2008optimum, murray2007coverage, yao2009can, konda2013optimal}. After the development of the \textit{unmanned aerial vehicle} (UAV), a considerable amount of research has focused on the placement of UAVs \cite{zorbas2016optimal} to obtain a solution for maximizing the area covered by UAVs while considering energy consumption, which is a critical metric to UAVs. Some other researchers studied tasks similar to depth cameras placement. \cite{olague2002optimal} provides an approach to the minimal geometry reconstruction error for an object whereas \cite{roy2004active} mentions the method to optimize the camera placement for object recognition. However, they are not suitable for larger-scale complex objects (such as rooms) and require the premeasurement of the target object.

Although many OCP solutions in various scenarios have been listed above, they are unsuitable for our task because of their specialization. We expect to apply OCP to the depth observation of indoor scenes. The working principles of different types of depth cameras are different; however, all depth cameras require the high-precision perception of the observation target, which limits their effective range. Depth cameras designed for indoor scenes can only guarantee the completeness and accuracy of the results within a few meters. This limitation makes depth cameras more dependent on suitable camera placement than traditional optical cameras. Further, the use of existing OCP frameworks to plan the placement of depth cameras may be insufficient because only a few of the existing OCP frameworks are designed to ensure the quality of the depth observations. Besides, most existing OCP algorithms require prior knowledge of scene information as input, such as floor plans or pre-measured three-dimensional data, which prevents them from online usage.

The main contributions of this study are summarized as follows:

\begin{itemize}
\item Proposed a reinforcement learning (RL) based camera placement system specialized for depth observation in indoor scenes.
\item Designed a sufficiently fast simulation environment required by training.
\item Designed an agent network for learning the placement of depth cameras; this network can deal with unstructured point cloud data.
\item Conducted comparative experiments with traditional offline approaches, which shows that the proposed system as an online algorithm can achieve more accurate depth observations than offline algorithms.
\end{itemize}

The rest of this paper is organized as follows. Section~\ref{chap:def} provides the detailed definition of the research problem. Section~\ref{chap:system} presents an overview and the detailed design of the proposed system from two parts of the system: the simulation environment and agent network. Subsequently, Section~\ref{chap:exper} describes the conducted experiments and the composition of the used dataset. Finally, the paper is concluded in Section~\ref{chap:concl}.

\section{Problem Definition}\label{chap:def}

For simplicity, we assume that all cameras are omnidirectional, and therefore, we use only camera positions as variable camera parameters. Omnidirectional depth cameras are already widely used in depth observation research, such as \cite{huang20176, luo2018parallax360, serrano2019motion}. Section~\ref{chap:sm} shows that our proposed system can be easily transferred to regular non-omnidirectional depth cameras. Another assumption considered for the sake of simplicity is that the number of cameras is fixed during the optimization process.

Based on these assumptions, the camera settings to be optimized include the camera positions. Therefore, the target of the depth camera placement for depth observation in indoor scenes can be given as finding the camera positions $\mathbf{p}$ as
\begin{equation}\label{eq:odcp}
\argmin_{\mathbf{p}\in\mathbb{S}}D(\mathbf{p}\,|\,\mathbf{C}),
\end{equation}
where $\mathbb{S}$ denotes the valid value range of $\mathbf{p}$, i.e., the whole indoor space; $\mathbf{C}$ is the point cloud data that represents the scene; $D(\mathbf{p}\,|\,\mathbf{C})$ is the difference between the ground truth and depth images observed by cameras placed as $\mathbf{p}$, and the research target is to minimize this difference.

The difference $D_\mathbf{j}(\mathbf{p}\,|\,\mathbf{C})$ between the depth camera observation and the ground truth at a given viewpoint $\mathbf{j}$ can be intuitively defined as the sum of pixel-wise differences on the depth image as
\begin{equation}\label{eq:odcp2}
D_\mathbf{j}(\mathbf{p}\,|\,\mathbf{C})=\sum_{pixel\in\mathcal{I}_\mathbf{j}}||\,d_{pixel}(\mathbf{p}\,|\,\mathbf{C}) - d^{'}_{pixel}(\mathbf{C})\,||,
\end{equation}
where $\mathcal{I}_\mathbf{j}$ denotes the depth image observed at $\mathbf{j}$; $d_{pixel}(\mathbf{p}\,|\,\mathbf{C})$ represents the pixel value on the depth image calculated with the partial point cloud observed by cameras at $\mathbf{p}$, and $d^{'}_{pixel}(\mathbf{C})$ represents the pixel value on the depth image calculated with the entire point cloud, i.e., the ground truth.

Ideally, $D(\mathbf{p}\,|\,\mathbf{C})$ should be defined as the sum of $D_\mathbf{j}(\mathbf{p}\,|\,\mathbf{C})$ for all viewpoints $\mathbf{j}$ in $\mathbb{S}$. However, this definition is difficult to calculate in practice. Therefore, the appropriate approach is to consider some sampling points in $\mathbb{S}$, and define $D(\mathbf{p}\,|\,\mathbf{C})$ as the sum of $D_\mathbf{j}(\mathbf{p}\,|\,\mathbf{C})$ on these sampling points. With this definition, the target can be formulated as
\begin{equation}\label{eq:odcp3}
D(\mathbf{p}\,|\,\mathbf{C})=\sum_{\mathbf{j}\in\mathbf{V}}D_\mathbf{j}(\mathbf{p}\,|\,\mathbf{C}),
\end{equation}
where $\mathbf{V}$ represents the set of sampling points.

\section{Reinforcement Learning System}\label{chap:system}

\subsection{System Overview}

\begin{figure}[t!]
\centering
\includegraphics[width=8.5cm]{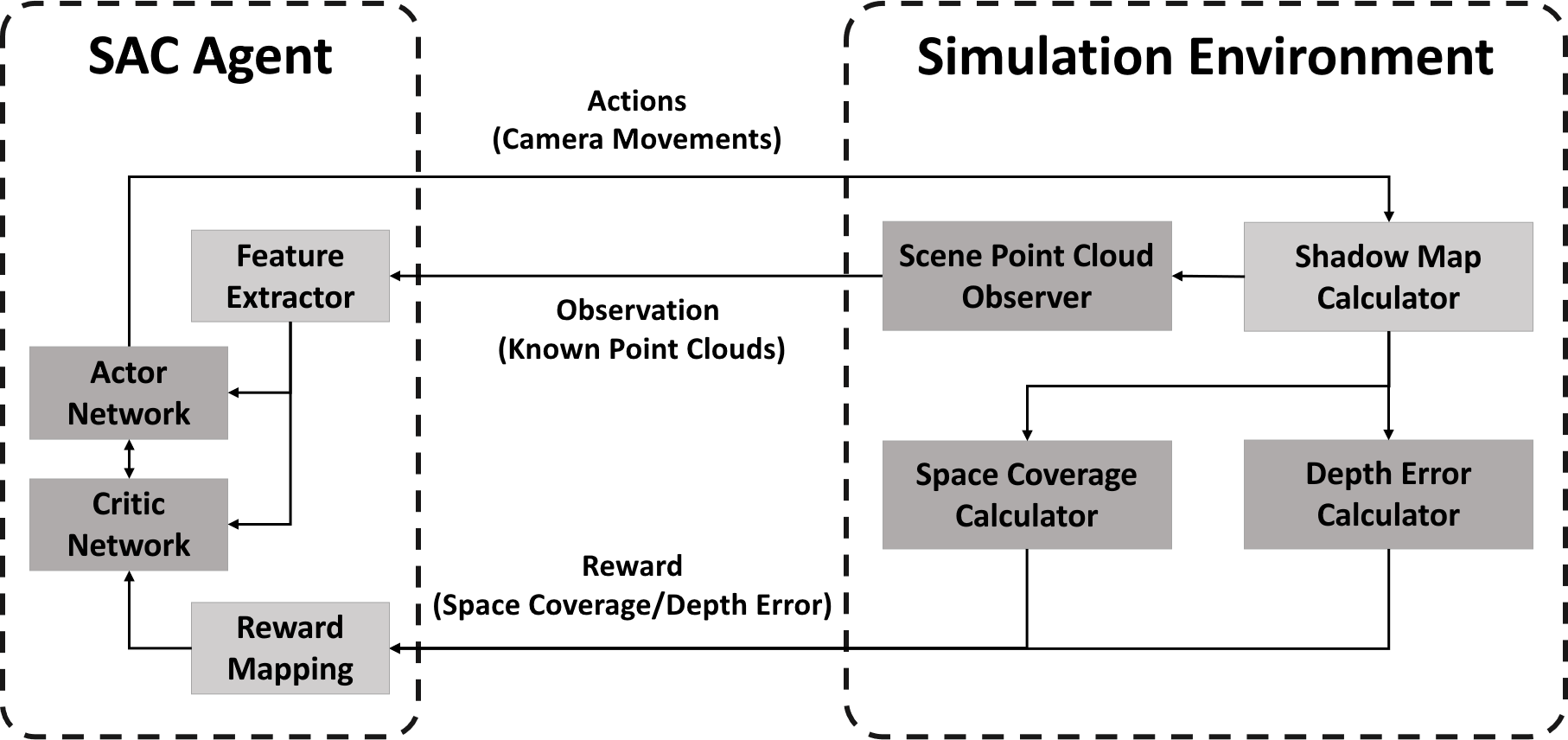}
\caption{Overview of training routine.}
\label{fig:train_ov}
\end{figure}

The overview of the training routine of the proposed system is illustrated in \figurename~\ref{fig:train_ov}. Similar to other reinforcement learning tasks, the training process of this system is conducted through an iterative interaction between the agent and the environment.

The \textit{soft actor critic} (SAC) \cite{haarnoja2018soft} algorithm is adopted as the agent backbone. The agent network comprises three parts: an actor network representing the strategy responsible for generating actions from the extracted features and estimated Q-value; a critic network responsible for the estimation of Q-value from the extracted features and action; and a feature extractor that processes point cloud data. The actor network and critic network use the implementation provided by the \textit{stable baselines3} \cite{stable-baselines3}, and therefore, they are not described in detail in this paper.

The actions produced by the agent network are designed to be camera movements in each turn. The agent generates actions for all camera globally. Given the actions output by the agent, the simulation environment becomes responsible for outputting the corresponding observation and reward. For our research, the output is the partial point cloud observed until the current turn and the reward of the movements (including the difference of space coverage and depth observation error compared to the last turn).

The evaluation routines are similar to the training routines that involve repeated interactions between the agent and the environment. However, the estimation of the Q-value and computation of the rewards on which it depends are unnecessary because there is no need to consider backpropagation. The modules related to them can be discarded from \figurename~\ref{fig:train_ov}. Moreover, when used in real-world applications, the entire simulated environment is replaced by the natural environment and only the agent network is required.

The camera movements generated by the agent gradually converge to zero after a certain number of interaction cycles. Thus, the environment is set with the maximum number of interactions. The loop is stopped after reaching this maximum value. The camera positions at the end are used as the final positions to evaluate the performance of the depth camera placement using (\ref{eq:odcp3}).

\subsection{Simulation Environment}

For our research, the required simulation environment must perform the following functions while considering both speed and accuracy:

\begin{itemize}
\item Scene point cloud observation 
\item Space coverage estimation
\item Depth observation error estimation
\end{itemize}

These functions correspond to the three dark-colored modules in the simulation environment in \figurename~\ref{fig:train_ov}. The proposed system implements these functions based on \textit{shadow maps}, which correspond to the light-colored module in the simulated environment in \figurename~\ref{fig:train_ov}.

\subsubsection{Shadow Maps Calculator}\label{chap:sm}

Basic shadow maps are similar to depth images because the depth information along different directions is stored pixel-by-pixel. The name \textit{shadow maps} is based on its application in real-time rendering. Shadow maps can be used to store distances from the nearest obstacle to the light source. Further, shadow maps can be used to evaluate whether the light source illuminates a specific fragment by comparing their distance with the shadow map value.

\begin{figure}[!t]
\includegraphics[width=8.5cm]{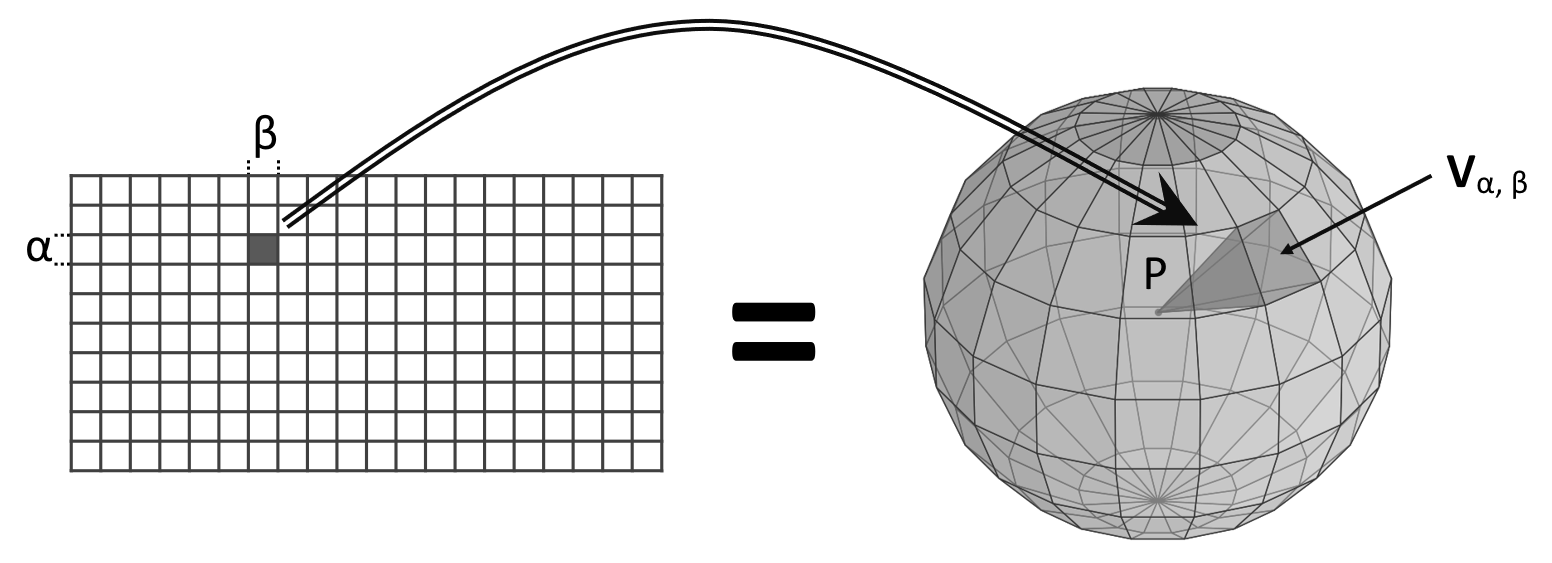}
\centering
\caption{Shadow maps for omnidirectional cameras. The left part represents the shadow map, and the right represents the corresponding coverage volume in the three-dimensional space of the painted one element in the left. $\alpha$ and $\beta$ are the indices of the elements in the shadow map and their corresponding spherical coordinate values. $\mathbf{P}$ is the camera position. The radius of the ball on the right is the effective range of the depth camera. $\mathbf{V}_{\alpha,\,\beta}$ is a quadrangular pyramid-like body surrounded by four radii (corresponding to the four vertices of the painted cell in the left image) and the spherical surface. The value of the painted cell is the minimum distance from all points in $\mathbf{V}_{\alpha,\,\beta}$ to the center of the sphere.}
\label{fig:sm}
\end{figure}

Shadow maps can be calculated directly from the point cloud data. Further, shadow maps can be defined in a spherical coordinate system with the camera as the center and independent of the radial distance because users are assumed to use omnidirectional cameras. The definition and calculation of the shadow map are shown in \figurename~\ref{fig:sm} and (\ref{eq:sm_sim_def}).
\begin{equation}\label{eq:sm_sim_def}
\textbf{SM}_i[\alpha,\,\beta] = \min_{\mathbf{point}\in\mathbf{V}_{i,\,\alpha,\,\beta}}||\,\mathbf{point} - \mathbf{P_i}\,||_2
\end{equation}
The above shadow maps are defined for omnidirectional depth cameras; however, they can still be used for normal (non-omnidirectional) depth cameras. The calculation is also easy for normal scenarios; only slight modifications (\ref{eq:sm_sim_def}) as required.
\begin{equation}\label{eq:sm_no_def}
\textbf{SM}_i[\alpha,\,\beta] = \min_{\mathbf{point}\in\mathbf{V}_{i,\,\alpha,\,\beta}\cap \mathbf{F_i}}||\,\mathbf{point} - \mathbf{P_i}\,||_2,
\end{equation}
where $\mathbf{F_i}$ denotes the viewing cone of the i-th camera. This change is essentially as an additional judgment. The shadow map is updated only when the point falls within the viewing cone of the camera.

Each point is considered a rectangle when calculating the shadow maps. Not only the cell in which the point falls but also the surrounding cells are updated. The rectangles are centered on the cell where the point is located; however, their size is not fixed. Each point is considered a cube whose side length is twice the distance from the point to the nearest neighbor in the point cloud. Without this amendment, the gap between the points may uncover points that should not be visible behind them.

\subsubsection{Visibility Test in Following Modules}

It is simple to conduct a visibility analysis to determine if a point is visible to a camera once the shadow map is calculated. The approach is to identify the cell where the point falls in the shadow map of the camera and to determine if the distance to that camera is less than the value stored in the cell. A small compensation is added to the shadow map value. Thus, the visible area is slightly expanded, and it covers visible points that would have been erroneously eliminated because of the sampling accuracy.

\subsubsection{Scene Point Cloud Observer}

One step to obtain the partial point cloud observed in the current turn is checking the visibility of all cameras for each point in the point cloud. If a point is visible by any camera, it is added to the list of observed points. Finally, this list is output as the simulation of the observation results. The observation flag is set to false each time the environment is reset. Once the flag of some point is set to true, it will always be true, which indicates that this point will be included in the observation.

\subsubsection{Space Coverage Calculator}

The space is discretized to estimate the space coverage. The bounding box of the scene point cloud is considered as the entire space, and it is then uniformly divided into small cells. Each cell in the grid determines whether it is in the visible area by checking the visibility of its center point. The number of cells that pass the visibility check is then multiplied by the volume of an individual cell to obtain the space coverage.

\subsubsection{Depth Error Calculator}

Calculating the depth error at a single point is straightforward. Owing to the similarity of shadow maps and depth images, they can be directly considered the depth information observed at a certain point. Therefore, to compare the depth observation obtained by placed cameras and the ground truth, shadow maps are calculated with the partial point cloud visible to placed cameras and the entire point cloud, respectively. Then, the depth error is calculated by accumulating the differences of the depth values pixel-by-pixel.

Nevertheless, it is difficult to compute the depth error for the entire space. The depth errors at some sampling points are used to represent the overall depth error. Then, the summation of the depth errors at all sampling points can estimate the overall depth error. Sampling points are placed in the center and four corners ($\frac{2}{3}$ of the position from the center point to the vertex of the bounding box) of the bounding box of the scene. The height of the sampling points is half of the height of the scene bounding box.

\subsection{Agent Network}

The critical parts of the agent network include a feature extractor module and a reward mapping module, excluding the actor and critic networks, which are not our contributions. The feature extractor module extracts latent features from the input observed point clouds. The reward mapping module preprocesses the rewards from the environment to guide the training direction better.

\subsubsection{Feature Extractor}

Inspired by other networks that extract features from point clouds \cite{qi2017pointnet, qi2017pointnet++}, the proposed feature extractor extracts point cloud features level-by-level through sampling and grouping. The core part of this module is the extraction unit, which extracts a set of key points and features from the input point cloud. The extraction unit is repeated three times in the feature extractor to extract the features of the different levels. The last layer is set to have only one key point, which extracts the global features of the point cloud. The features extracted from the point cloud are concatenated with manually extracted features (bounding box of observed point cloud and current camera positions) as the input of the subsequent actor and critic network.

\begin{figure}[t!]
\centering
\includegraphics[width=8.5cm]{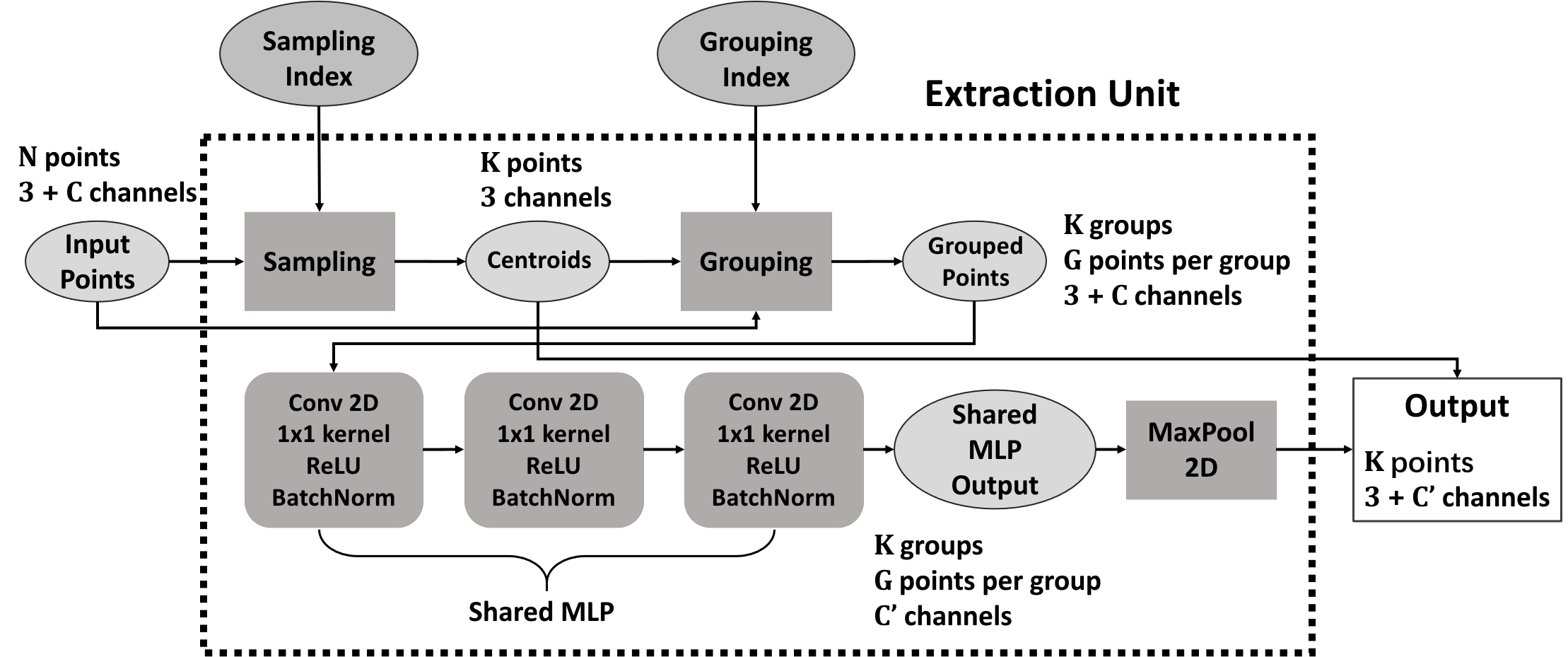}
\caption{Forward data flow diagram of the extraction unit in feature extractor module.}
\label{fig:df_feu}
\end{figure}

The specific structure of the extraction unit is shown in \figurename~\ref{fig:df_feu}. The input point cloud is first sampled to obtain $\mathbf{K}$ key points, and then, the point clouds are grouped with these key points as the center points. Each group has $\mathbf{G}$ points. Sampling and grouping operations are the same as in \cite{qi2017pointnet++}. These operations select some points from the point cloud based on indices, which are constant for each point cloud. Therefore, these indices are precomputed in the observation cache module and passed into the extraction unit when extracting features.

Then, the grouped points are passed through a multilayer perceptron (MLP) with shared weights to extract latent features. The MLP comprises three layers with the same output channel number $\mathbf{C'}$. The following maximum pooling layer filters out the maximum components of each group in the output of the MLP. The filtered components are concatenated with the center point position of each group. Therefore, the final output of the feature extraction unit is similar to a point cloud containing $\mathbf{K}$ points, and each point has 3 position channels and $\mathbf{C'}$ feature channels.

\subsubsection{Reward Mapping}

The simulation environment provides two reward outputs: space coverage $\textit{SC}$ and depth observation error $\textit{DOE}$. Further, a penalty term $\textit{P}$ for preventing the actions of moving the camera outside the room is added. This term is a Boolean variable, and it is calculated whenever the rewards are calculated. When any camera position is outside the bounding box of the scene, $\textit{P}$ is set to 1. Otherwise, it is set to 0. These three reward terms are combined into a reward scalar $\textit{Rew}$ as
\begin{equation}\label{eq:rew1}
\begin{split}
\textit{Rew}\,&(\,\textit{SC},\,\textit{DOE},\,\textit{P}) =\\
&(K_{\textit{SC}} \times\Delta\textit{SC} + K_{\textit{DOE}} \times\Delta\textit{DOE})\times(1 - \textit{P})\\
&+ K_\textit{P}\times\textit{P},
\end{split}
\end{equation}
where $\Delta\textit{SC}$ and $\Delta\textit{DOE}$ represent the difference in space coverage and depth observation error compared to the previous turn. $K_{SC}$, $K_{DOE}$, and $K_P$ denote the weights of space coverage, depth observation error, and penalty terms, respectively.

When using (\ref{eq:rew1}) directly for the loss calculation of the Q-network, the agent rapidly expands the coverage area of the camera at the beginning, and this is accompanied by a significant decrease in the depth error. Then, the agent may appear sluggish. The agent keeps the cameras immobile even though more suitable (lower depth error) positions may exist near the cameras.

For this problem, we propose a solution that involves mapping rewards by functions with sufficiently large positive first-order derivatives and positive second-order derivatives. In the case of already large rewards, sufficiently large positive first-order derivatives allow even small increments of rewards to produce significant increments of mapped rewards. Furthermore, the positive second-order derivative makes this effect increasingly enhanced as the base reward becomes larger. In this specific implementation, we use the third power function as the mapping function.

\section{Experiments}\label{chap:exper}

The experiments described in this paper are performed on an Ubuntu 18.04 server with two \textit{AMD Ryzen Threadripper 3970X} CPUs, 128GB RAM, and one \textit{NVIDIA TITAN RTX} graphics card.

\subsection{Dataset}

Our training and evaluating datasets are based on the S3DIS dataset \cite{armeni_cvpr16}, which is a large-scale indoor scene dataset obtained by scanning the real-world campus rooms by depth cameras, including 3 buildings, 6 areas, and a total of 272 scenes. The point cloud data of scenes from area 1 to area 5 are used to construct the training dataset, and data from area 6 are used to construct the evaluation dataset.

However, not all scenes in the S3DIS dataset are suitable for the experiments. 10 scenes from each area are selected based on the following rules to form the experiment datasets.

\begin{itemize}
\item \textbf{Scene shape completeness:}
The selected scene must have closed walls and ceilings. An open space may interfere with the measurement of depth observation errors and evaluation of whether the camera is inside the scene. This rule only applies to the training dataset. In order not to introduce bias to the evaluation set, the evaluation set still contains open scenes.
\item \textbf{Scene size:}
The selected scene must be of the appropriate size to be roughly covered by the chosen number of cameras without too much omission or overlap. If the scene is too small, then the camera can easily cover the whole scene. If the scene is too large, a good depth observation cannot be obtained no matter how the cameras are placed because the number of cameras is fixed. Optimizing camera placement in both cases is not meaningful, and therefore, they cannot be used for our experiments.
\item \textbf{Flat room:}
We restrict camera positions in experiments on a plane parallel to the ground to simplify the calculation. This restriction makes the simplified model handle flat rooms. Therefore, we excluded non-flat rooms such as stairwells from the training and evaluation sets.
\end{itemize}

Further, segmented scenes in the S3DIS dataset are all axis-aligned. To avoid this property from introducing bias into datasets, two additional scenes for each scene are added by rotating the scene by 60\degree~ and 120\degree~ around the center of its bounding box. Thus, the final dataset used for the experiments contains $60\times 3 = 180$ scenes, 150 of which are used for training and 30 for evaluation.

\subsection{Comparative Experiment}

\subsubsection{Baselines}

Two versatile non-deep-learning OCP solutions, TDSA \cite{liu2013optimal} and BPSO \cite{morsly2011particle} are used as baselines to compare with our proposed system.

\subsubsection{Metrics}

The primary metric of the comparative experiment is the depth observation error, i.e., (\ref{eq:odcp3}), which is calculated in the simulation environment. Therefore, we can directly use the depth observation error output from the simulation environment as the evaluation metric.

\subsubsection{Results}

\begin{figure}[t!]
\centering
\includegraphics[width=8.5cm]{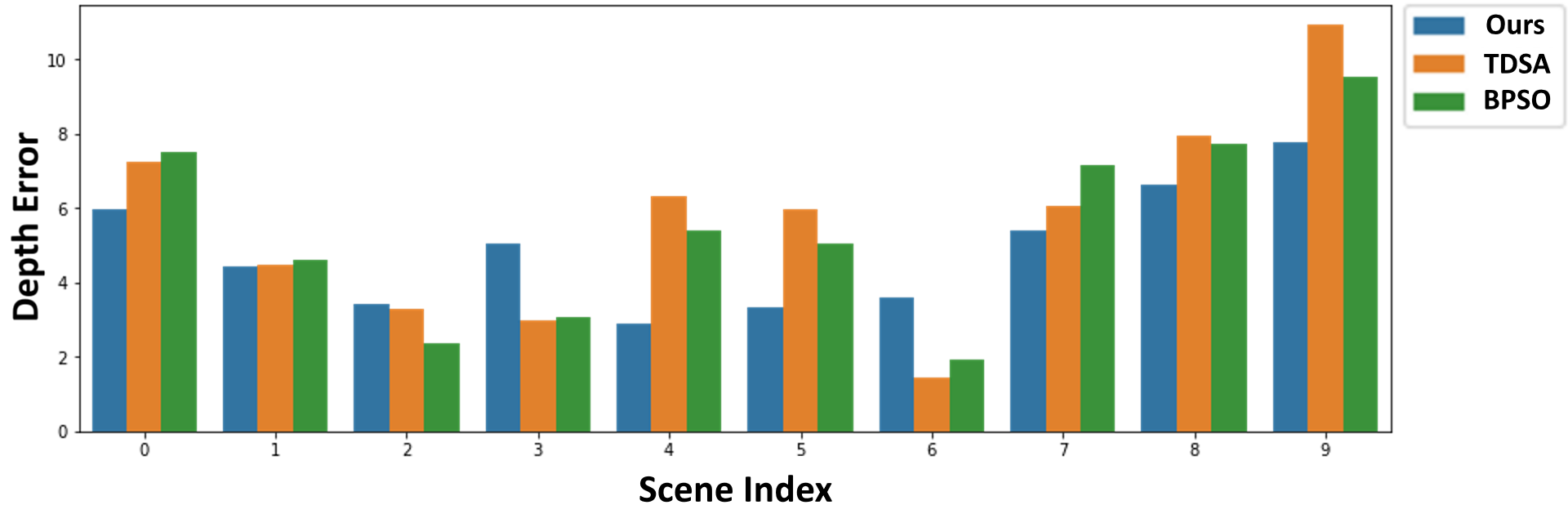}
\caption{Depth observation error of each approach in each evaluation scene.}
\label{fig:result}
\end{figure}

\begin{table}[!t]
\renewcommand{\arraystretch}{1.3}
\caption{Depth error sum of all evaluation scenes}
\label{table:error_sum}
\centering
\begin{tabular}{cccc}
\hline
Approach & Ours & TDSA & BPSO \\
\hline
Depth Error Sum & \textbf{48.4} & 56.6 & 54.3 \\
\hline
\end{tabular}
\end{table}

\begin{table}[!t]
\caption{Computation time for each approach}
\label{table:time}
\centering
\begin{tabular}{cccc}
\hline
Approach & Ours (Training) & TDSA & BPSO \\
\hline
Time & $\sim$ 9 h & $\sim$ 6 h & $\sim$ 13 h \\
\hline
\end{tabular}
\end{table}

\figurename~\ref{fig:result} shows the depth observation error obtained by each approach in each evaluation scene. The error of each scene is the average of the depth observation errors obtained by itself and those in the two derived scenes. \tablename~\ref{table:error_sum} shows the sum of errors for all scenes. The proposed system performs better in 7/10 scenes, and the total error is the smallest. This result shows that our system is better than general algorithms on camera placement for depth observation tasks. \tablename~\ref{table:time} summarizes the computation times for each approach. The time required by the baselines is the computation time for all 30 evaluation scenes. The results indicate that even if our approach is considered an offline algorithm, its time consumption remains acceptable.

\subsection{Training Statistics}

\begin{figure*}[!t]
\centering
\subfloat[Performance]{\includegraphics[width=2.7in]{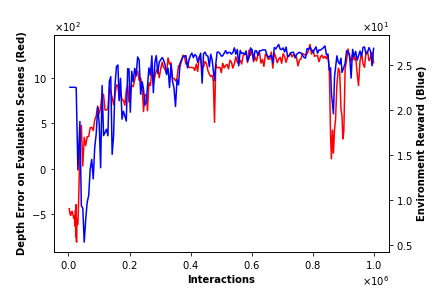}}
\hfil
\subfloat[Losses]{\includegraphics[width=2.7in]{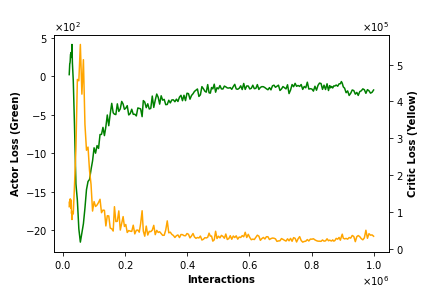}}
\caption{Performance and losses of agent measured during training.}
\label{fig:train_stat}
\end{figure*}

\figurename~\ref{fig:train_stat} shows the additional details of the training process. \figurename~\ref{fig:train_stat} shows that after one million interactions, the performance of the agent has almost stabilized, and the losses of modules gradually converged.

\subsection{Example Visualization}

\begin{figure*}[!t]
\centering
\subfloat{\includegraphics[height=1.5in]{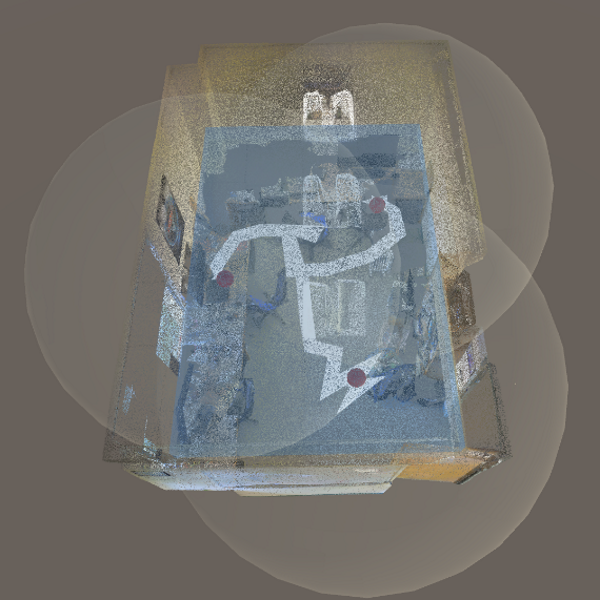}}
\hfil{\includegraphics[height=1.5in]{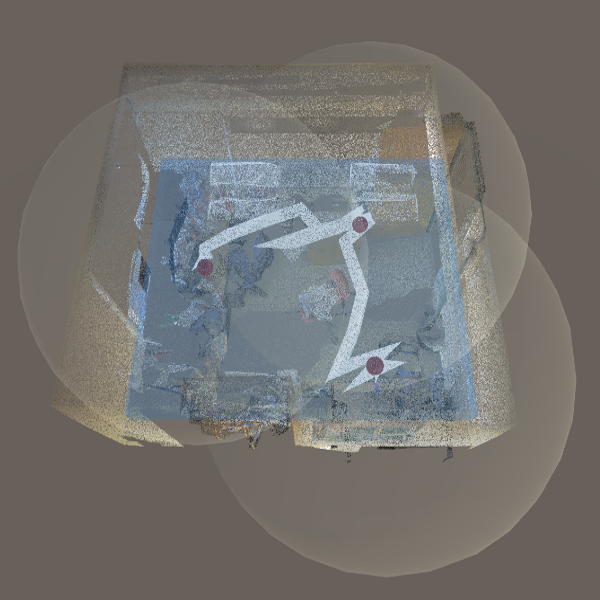}}
\hfil
\subfloat{\includegraphics[height=1.5in]{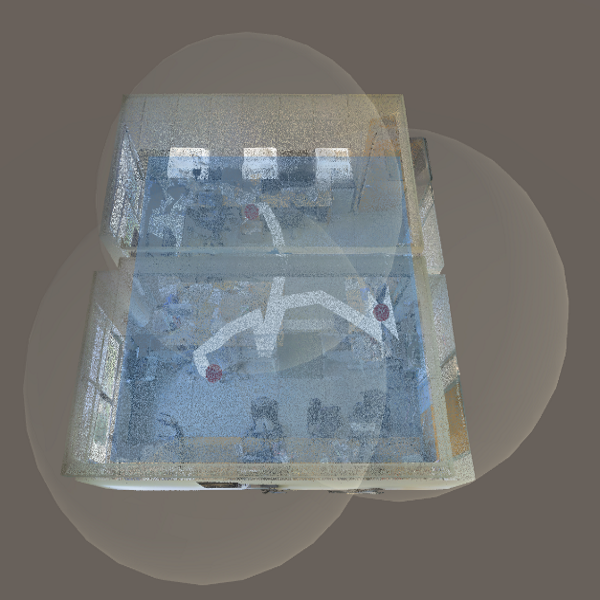}}
\caption{Examples of camera placement results and trajectories output by the proposed system.}
\label{fig:example}
\end{figure*}

\figurename~\ref{fig:example} shows some examples of camera placement results and trajectories output by the proposed system in the test scenes. The red balls, white lines, and semi-transparent white balls represent the final camera positions output by the system, trajectories of the camera movement during the repeated interactions with the environment, and influential ranges of the cameras, respectively.

\section{Conclusion}\label{chap:concl} 
 
In this study, a reinforcement learning-based OCP system for depth observation tasks in indoor scenes was proposed. As an online algorithm, it behaves better in obtaining more accurate depth observations compared to offline baselines. Therefore, it is more competent for depth camera placement in scenarios where there is no prior knowledge of the scenes or where a lower depth observation error is the main objective.

One of the main limitations of the proposed system is that it can be overly constrained to some corners of the scene in some scenes, thus neglecting other regions. This is the main factor why the system did not beat the baselines in all test scenes. Another major limitation is that it is restricted to using a fixed number of depth cameras. Therefore, future work may include introducing collaborative learning and exploring a way to encode the movements of variable-length cameras.

% \section*{Acknowledgment}

\bibliographystyle{IEEEtran}
\bibliography{IEEEabrv, ref}

\end{document}